\documentclass[runningheads]{llncs}

\usepackage{eccv}

\usepackage{eccvabbrv}

\usepackage{graphicx}
\usepackage{booktabs}

\usepackage{bbding}
\usepackage{amsmath}
\usepackage{multirow}
\usepackage{xcolor}
\usepackage{colortbl}
\usepackage{dsfont}

\usepackage{wrapfig}

\definecolor{grey}{rgb}{0.32, 0.32, 0.32}

\usepackage[accsupp]{axessibility}  

\usepackage[pagebackref,breaklinks,colorlinks,citecolor=eccvblue]{hyperref}

\usepackage{orcidlink}

\begin{document}

\title{Unlocking Textual and Visual Wisdom: Open-Vocabulary 3D Object Detection Enhanced by Comprehensive Guidance from Text and Image} 

\titlerunning{Unlocking Textual and Visual Wisdom}
\author{Pengkun Jiao\inst{1,2} \and
Na Zhao\inst{3}\thanks{Corresponding Author}  \and
Jingjing Chen\inst{1,2} \and
Yu-Gang Jiang \inst{1,2}
}

\authorrunning{P.~Jiao et al.}

\institute{Shanghai Key Lab of Intell. Info. Processing, School of CS, Fudan University \and
Shanghai Collaborative Innovation Center on Intelligent Visual Computing
\and
Singapore University of Technology and Design \\
\email{pkjiao23@m.fudan.edu.cn}, 
\email{na\_zhao@sutd.edu.sg},
\email{\{chenjingjing,ygj\}@fudan.edu.cn} 
}

\maketitle

\begin{abstract}
Open-vocabulary 3D object detection (OV-3DDet) aims to localize and recognize both seen and previously unseen object categories within any new 3D scene. While language and vision foundation models have achieved success in handling various open-vocabulary tasks with abundant training data, OV-3DDet faces a significant challenge due to the limited availability of training data. Although some pioneering efforts have integrated vision-language models (VLM) knowledge into OV-3DDet learning, the full potential of these foundational models has yet to be fully exploited.
In this paper, we unlock the textual and visual wisdom to tackle the open-vocabulary 3D detection task by leveraging the language and vision foundation models. We leverage a vision foundation model to provide image-wise guidance for discovering novel classes in 3D scenes. Specifically, we utilize a object detection vision foundation model to enable the zero-shot discovery of objects in images, which serves as the initial seeds and filtering guidance to identify novel 3D objects. Additionally, to align the 3D space with the powerful vision-language space, we introduce a hierarchical alignment approach, where the 3D feature space is aligned with the vision-language feature space using a pretrained VLM at the instance, category, and scene levels. Through extensive experimentation, we demonstrate significant improvements in accuracy and generalization, highlighting the potential of foundation models in advancing open-vocabulary 3D object detection in real-world scenarios.

\end{abstract}

\section{Introduction}
3D object detection serves as a fundamental component in understanding 3D scenes, playing a pivotal role in various applications \cite{arnold2019survey_on_auto_drive,mao20233d_od_for_autodrive,wu2023transformation_3det, jiao2023msmdfusion} such as autonomous driving and robot interaction. However, conventional approaches to 3D object detection \cite{zhou2018voxelnet,misra20213detr,yin2021center, yang2018pixor, pan20213d, zhao2020sess, sheng2022rethinking, han2024dual} often operate under the assumption that the detection targets during testing remain consistent with those observed during training. This assumption fails to reflect the dynamic and evolving nature of real-world scenarios, where the objects within scenes can vary and expand over time. Consequently, the capability of open-vocabulary 3D object detection, which enables the localization and recognition of both seen and previously unseen objects within new scenes, becomes essential for their practical deployment in real-world settings.

To achieve open-vocabulary capacity, image-based methods \cite{yao2022detclip, li2021glip,lai2023lisa} leverage internet-scale image-text pairs to train a unified feature alignment space. In contrast to significant achievements in its 2D counterpart, open-vocabulary 3D object detection (OV-3DDet) \cite{cao2023coda,lu2023ov3det,zhu2023l3det} faces critical challenge due to the scarcity of training data, which impedes the 3D detection models from effectively acquiring the ability for open-vocabulary inference. Fortunately, the success of large language and vision foundation models \cite{radford2021clip,zhou2022detic,liu2023dino,lai2023lisa}, such as vision-language models (VLMs) and large language models (LLMs), holds promise for benefiting 3D open-vocabulary learning. Several previous works \cite{cao2023coda, zhang2022pointclip, zhu2022pointclip2, lu2023ov3det} have demonstrated the feasibility of leveraging VLMs by using images as a medium to align text and images with 3D space for open-vocabulary learning in the 3D domain. For example, OV3DET \cite{lu2023ov3det} employs Detic \cite{zhou2022detic} to generate 2D bounding boxes (bboxes), which are then back-projected to 3D space to generate 3D bboxes. It aligns the 3D-image-text feature space using CLIP \cite{yao2022detclip} through category-level contrastive learning. Another recent work, CoDA \cite{cao2023coda}, also leverages CLIP but to provide semantic priors for selecting novel 3D objects from the class-agnostic 3D object detector. CoDA aligns 3D features to image features at the instance level and 3D features to text features at the category level. Additionally, instead of using images as a medium, L3DET \cite{zhu2023l3det} directly augments 3D scenes by injecting novel objects from external object-level datasets into the scenes. It aligns 3D features to text features extracted from LLMs (\ie RoBERTa \cite{liu2019roberta}) via category-level contrastive learning.

Despite the efforts of prior works, they have not fully capitalized on foundational models or effectively integrated valuable 3D information with these models. For instance, L3DET overlooks the utilization of VLMs, which excel in zero-shot tasks and exhibit a smaller domain gap with 3D data compared to LLMs. Furthermore, its simplistic injection augmentation approach may lead to unrealistic scenes with incongruous contextual information, thereby undermining the effectiveness of 3D object detection. Similarly, CoDA only employs VLMs as a prior for filtering novel 3D objects, relying solely on a class-agnostic 3D object detector trained on available annotations for 3D object discovery. Consequently, the discovery of 3D novel objects is constrained by the 3D inputs, making it challenging to detect classes with small size, sparse density, or insignificant structures. On the other hand, OV3DET heavily relies on vision foundation model and overlooks the valuable 3D inputs, which could provide rich geometry clues for 3D object discovery. 
Moreover, these methods predominantly focus on feature alignment either at the instance level or category level, neglecting to align the feature space comprehensively. 
 
\begin{figure}[t]
\centering
\includegraphics[width=1\textwidth]{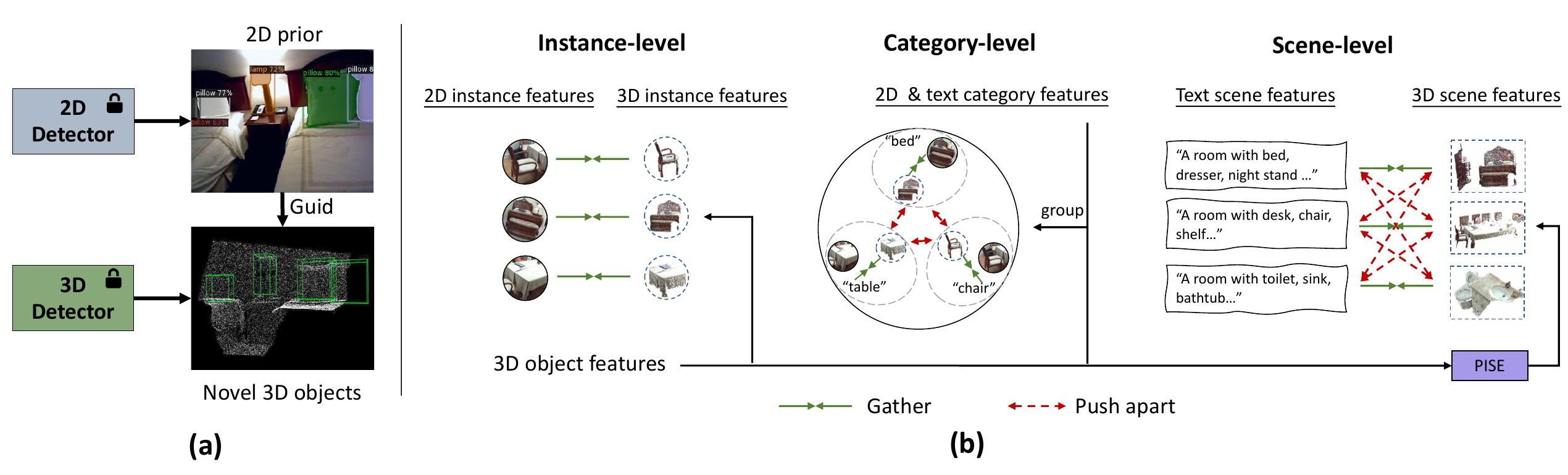} 
\vspace{-20pt}
\caption{\small{\textbf{Illustration of the two key components of our proposed INHA:}
(a) Image-guided novel object discovery (IGND) employs a vision foundation model to extract 2D bboxes and use them as prior to guide the discovery of 3D novel objects.
(b)  Hierarchical feature space alignment aligns the 3D feature space with the vision-language feature space at the instance, class, and scene levels.
}}
\label{fig:teaser} 
\end{figure}

To address these limitations, we propose a novel Image-guided Novel class discovery and Hierarchical feature space Alignment (INHA) approach. INHA leverages foundation models to unlock the full potential of text and image information for open-vocabulary 3D object detection. Figure \ref{fig:teaser} illustrates the two key components of INHA. In image-guided novel object discovery (Figure \ref{fig:teaser}a), we harness the power of vision foundation models to search and select 3D novel objects. Specifically, we use an off-the-shelf open-vocabulary 2D detector to locate 2D objects in the image. For detected 2D objects, we utilize their centroids as initial seeds to generate additional 3D object proposals and leverage their 2D bounding boxes to select reliable novel objects. This approach leads to improvement in the recall of 3D novel object discovery, as demonstrated in Figure~\ref{img:training_iter}. The discovered novel objects are combined with the given base objects to retrain the 3D detector, enhancing its class-agnostic 3D detection capability. Additionally, we introduce a hierarchical feature space alignment mechanism, aligning the 3D feature space with the vision-language feature space at instance-level, category-level, and scene-level, as shown in Figure \ref{fig:teaser}b. 
The incorporation of scene-level alignment is to capture the occurrence relation of classes across modalities. Since each scene contains a set of objects, direct comparison of this set to a text description of the scene is feasible. To accomplish this, we design a Permutation-Invariant Scene feature Extraction (PISE) module to extract 3D scene features and align them with the embedding of the scene description text.


\vspace{1pt}
Our main contributions can be summarized as follows:
\begin{itemize}
\item We propose a novel framework named INHA that exploits comprehensive guidance from text and images through language and vision foundation models, enhancing 3D open-vocabulary learning capacity. 

\vspace{2pt}
\item We introduce an image-guided novel object discovery (IGND) mechanism to effectively integrate valuable 3D information with image information from vision foundation models, facilitating the discovery of more 3D novel objects.

\vspace{2pt}
\item We design a permutation-invariant scene feature extraction (PISE) module to encode class occurrence relations in a scene. Additionally, we present hierarchical alignment of the 3D feature space with the vision-language feature space at instance, category, and scene levels.

\vspace{2pt}
\item Extensive experiments demonstrate the effectiveness of our proposed method. Our method achieves state-of-the-art performance in open-vocabulary 3D object detection on two benchmark datasets, SUN RGB-D and ScanNetv2.
\end{itemize}


\section{Related Work}
\noindent\textbf{3D Object Detection} 
endeavors to locate and identify 3D objects within a given scene, with numerous approaches proposed to address this challenge. 
VoteNet \cite{qi2019votenet} introduced a point voting strategy, leveraging PointNet$++$\cite{qi2017pointnetplusplus} for processing 3D points.
Evolving from VoteNet, MLCVNet \cite{xie2020mlcvnet} introduces additional modules for point voting, aiming to capture multi-level contextual information and enhance overall detection performance.
The advent of transformer-based methods has significantly shaped the landscape of 3D object detection. Notably, GroupFree \cite{liu2021groupfree} utilizes a transformer as the prediction head, eliminating the need for manually crafted grouping and harnessing the transformer's capabilities to improve detection accuracy. Recently, the trend has shifted towards end-to-end models, with 3DETR \cite{misra20213detr} standing out as the pioneer in employing an end-to-end transformer architecture for 3D object detection. It utilizes bipartite graph matching to establish associations between predictions and ground truths. 
However, traditional 3D object detection methods only detect objects seen during the training stage and cannot handle the open-vocabulary scenario.


\noindent\textbf{Open-vocabulary Object Detection} aims to detect objects that include both seen and unseen classes in an image. Since the success of CLIP in associating language and the 2D domain, research on zero-shot learning and open-vocabulary has become a trend. For example, Detic \cite{zhou2022detic} uses ImageNet to train the classifiers of detectors, expanding the vocabulary of detectors to tens of thousands of concepts. GLIP \cite{li2021glip} and MDETR \cite{kamath2021mdetr} treat detection as the grounding task and adopt a text query for the input image to predict corresponding boxes.
GLIPv2 \cite{zeng2023clip2} reformulates the detection task, introduces a novel region-word level contrastive learning task, and includes masked language modeling. 
The success of open-vocabulary 2D detection methods provides the potential to facilitate OV-3DDet. Our method also leverages a pretrained language-vision model to provide guidance for 3D object detection.

\noindent \textbf{Open-vocabulary 3D Object Detection} (OV-3DDet) presents a substantial challenge, aiming to detect and locate objects in a 3D space, encompassing both known and unknown object categories. In OV3DET \cite{lu2023ov3det}, a 2D detector is employed to generate pseudo-labels. These labels are utilized for training a class-agnostic detector, followed by a contrastive learning step that aligns the 3D feature space with image and language features.
L3DET \cite{zhu2023l3det} contributes to the field by enriching 3D scene datasets through the introduction of novel objects and associated text descriptions. It leverages cross-domain, category-level contrastive learning to align feature spaces between point clouds and language, facilitating effective cross-modal reasoning.
CoDA \cite{cao2023coda} takes a distinctive approach by integrating 3D box geometry priors and 2D semantic open-vocabulary priors for novel object discovery. Discovered novel objects contribute to subsequent detector training, aligning the 3D feature space with the vision-language feature space through class-agnostic knowledge distillation and class-aware contrastive learning.
However, CoDA exhibits inaccuracies and insufficiencies in novel class discovery and feature alignment. While PLA \cite{ding2023pla} introduces alignment at the instance, category, and scene levels, its primary focus lies within the segmentation setting, resulting in a higher cost at the point-wise level rather than the object-wise level. In response to these challenges, we propose an alternative approach.

\section{Method}
\subsection{Problem Definition}
In OV-3DDet, we are given the point cloud of a scene, denoted as $P=\left \{\textbf{p}_{i} \in \mathcal{R} ^{3} \right \}$, as well as the associated image of the scene, denoted as $\textbf{I} \in \mathcal{R}^{3 \times H \times W} $.
For each point cloud, the objects present in the scene are denoted as $O_{3D} = \left \{(\textbf{B}_{n}, \text{c}_{n})\right \}_{n=1}^{N} $, 
where $\textbf{B}_{n} \in \mathcal{R}^{7}$ represents the 3D bounding box parameters, including the center, size, and heading angle, and $\text{c}_n \in C$ represents the category of the object.  
Among the objects in a scene, some are initially annotated with labels, denoted as base objects $O_{3D}^\mathcal{B}  = \left \{( \textbf{B}^{\mathcal{B} }_{j}, \text{c}^{\mathcal{B}}_{j} )\right \} $, $\text{c}^{\mathcal{B} }_{j}$ belongs to the base label space $C^{\mathcal{B}}$, while the remaining unlabeled objects are 3D novel objects $O^{\mathcal{N}}_{3D} = \left \{(\textbf{B}^{\mathcal{N}}_{k}, \text{c}^{\mathcal{N}}_{k})\right \} $, where $\text{c}_{k}^{\mathcal{N}}$ belongs to the novel label space $C^{\mathcal{N}}$, and $C^{\mathcal{B}} + C^{\mathcal{N}} = C$. Our objective is to train an open-vocabulary 3D object detector capable of localizing and recognizing both base and novel objects in any new point cloud.

\begin{figure*}[t]
\centering
\includegraphics[width=1\textwidth]{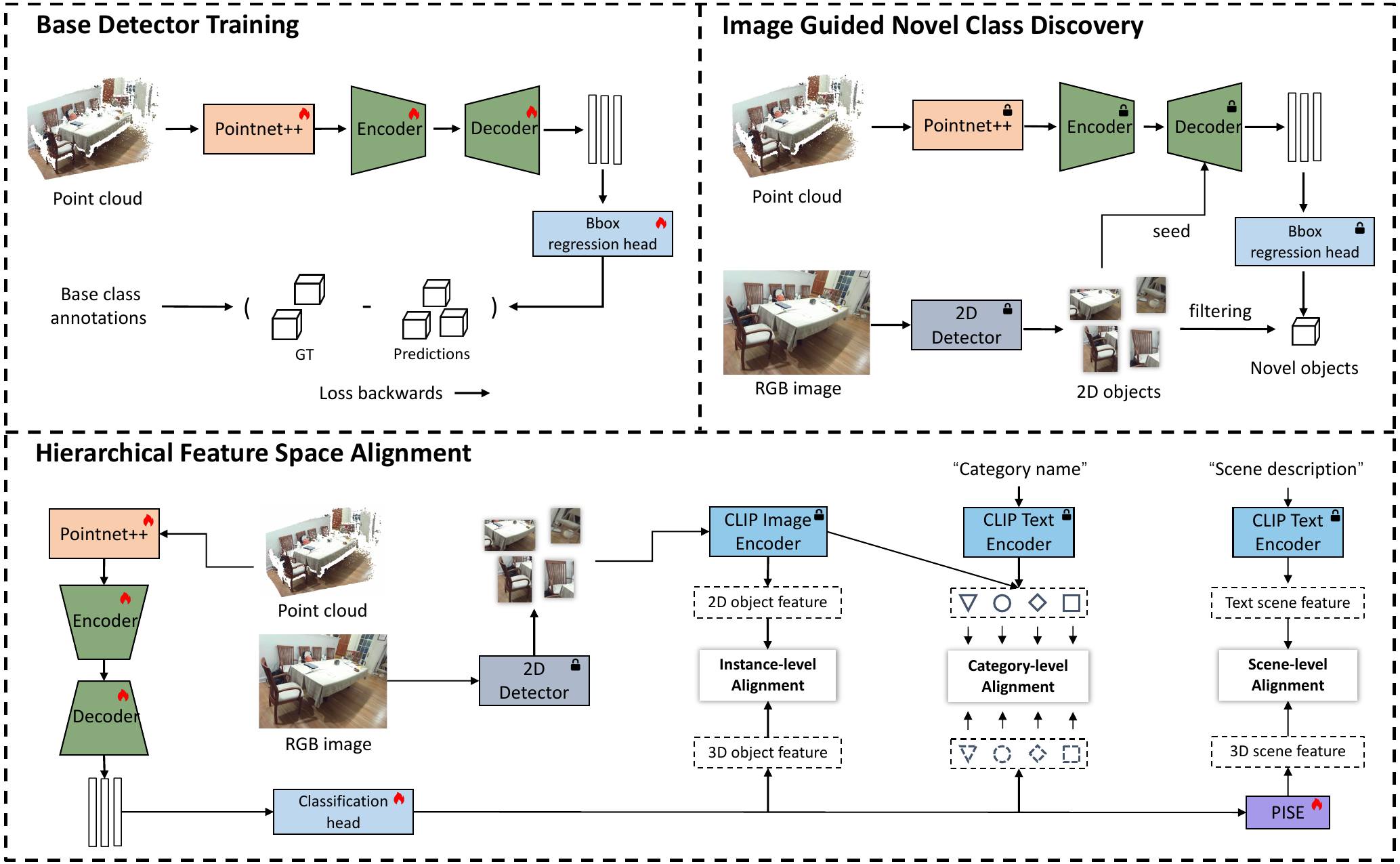} 
\vspace{-0.2in}
\caption{\small{\textbf{Illustration of our proposed INHA framework.} Our INHA framework primarily consists of three stages. Firstly, we train a base detector solely utilizing base objects. Subsequently, in the second stage, we enhance the 3D detector by incorporating discovered novel classes from the IGND module. Finally, in the third stage, we perform hierarchical alignment of the 3D feature space with the vision-language feature space at the instance, category, and scene levels.}
}
\label{img:framework}
\vspace{-0.2in}
\end{figure*}

\subsection{INHA Overview}
Our proposed Image-guided Novel class discovery and Hierarchical feature space Alignment (INHA) approach adopts 3DETR \cite{misra20213detr} as our 3D object detector, which comprises PointNet$++$, Transformer encoder, Transformer decoder, a bounding box regression head, and a classification head. The training process consists of three stages, illustrated in Figure \ref{img:framework}.
In the first stage, we train a base class-agnostic 3D object detector. More specifically, we remove the classification-related training and solely utilize bounding box regression loss $\mathcal{L}_{\text{box}}$ \cite{misra20213detr} for training the detector. 
After the first stage, the detector is capable of detecting objects in a point cloud without considering their class information. 
In the second stage, we incorporate both the 2D detector and the 3D detector to discover novel objects (\cf Sec.~\ref{sec:IGND}).
The discovered objects are stored and then used for re-training the 3D detector. This process enhances the 3D detector's ability to handle novel classes.
In the final stage, we introduce a hierarchical paradigm (\cf Sec.~\ref{sec:hierarchical_align}) that aligns the feature space at instance, class, and scene levels.

\begin{figure*}[t]
\centering
\includegraphics[width=1\textwidth]{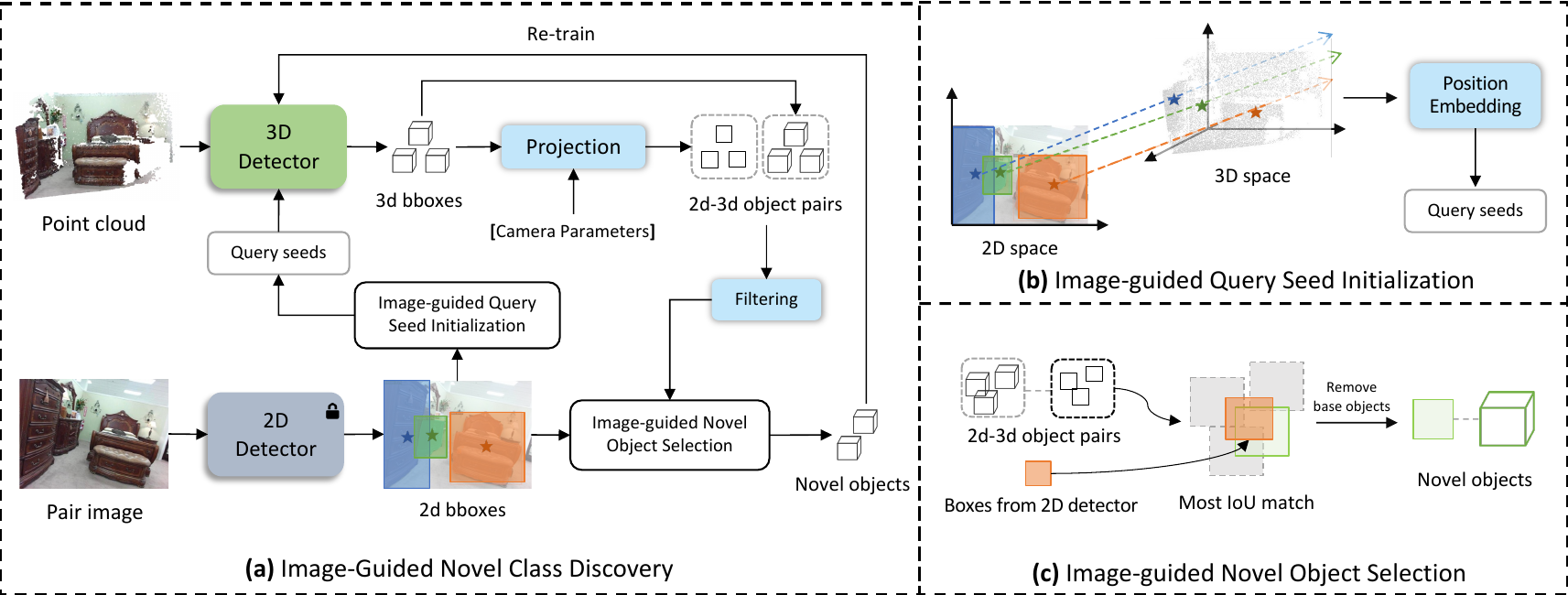} 
\vspace{-0.2in}
\caption{
\small{
\textbf{Illustration of IGND.} 
(a) The overarching framework of IGND integrates the functionalities of both 2D and 3D detectors for novel object discovery.
(b) In this step, the translation of 2D object centers into 3D space enriches the pool of query seeds, facilitating the generation of novel object proposals.
(c) Projected onto 2D boxes, 3D bounding boxes are matched with 2D detected objects based on their IoU scores to select the most suitable candidates.
}}
\label{img:igod}
\vspace{-0.1in}
\end{figure*}

\subsection{Image-Guided Novel Class Discovery}
\label{sec:IGND}
The image provides rich appearance cues that aid in identifying cluttered objects. Leveraging the capabilities of recent open-vocabulary 2D object detectors, even extremely small or occluded novel objects can be identified in images. For instance, a distant object can be easily discerned by a 2D detector using only a few pixels. In contrast, 3D object detectors struggle to recognize distant objects with very few points. However, 3D object detectors excel in capturing rich 3D geometrical information, resulting in more precise localization predictions. Recognizing the complementary nature of these modalities, we present the Image-Guided Novel Class Discovery (IGND) module, illustrated in Figure \ref{img:igod}, to discover more 3D novel objects. In this module, we utilize a pretrained open-vocabulary 2D detector, \ie Detic \cite{zhou2022detic}, to extract valuable object-level information (2D object bboxes) from images. This information is then effectively integrated with valuable 3D data to guide the discovery of 3D novel objects.  Specifically, this guidance occurs in two key steps: a) lifting the centroids of the 2D objects to 3D space to provide supplementary query seeds for generating more 3D object proposals, and b) utilizing the bounding boxes of the 2D objects to select reliable novel 3D bounding boxes. Through these two guided processes, we enhance the recall rates of novel classes, as depicted in Figure \ref{img:training_iter}.


\noindent \textbf{Image-guided Query Seed Initialization.}
The transformer-based 3D detector \cite{misra20213detr, lu2023ov3det} typically employs position encoding of multiple points as initial query seeds to propose 3D objects, which are conversely sampled by object-agnostic sampling algorithms, \eg random sampling or the farthest point sampling \cite{qi2017pointnetplusplus}. 
The quality of query seeds significantly influences the effectiveness of novel object proposals. Given that 2D detectors can reliably identify novel objects, we elevate the centers of 2D objects to 3D space to obtain supplementary query seeds.
Let $O_{\text{2D}}= \left \{ (\textbf{b}_{m}, \text{c}_{m}) \right \}_{m=1}^M$ denote the detected $M$ 2D objects in an image, where $\textbf{b}_{m} \in \mathcal{R}^4$ and $c_m \in C$. We lift the centers of 2D bounding boxes to 3D space.
These lifted 3D points are then encoded into position embeddings \cite{kamath2021mdetr, misra20213detr}  and added to query seeds, which can be used to facilitate novel object proposals within the Transformer decoder framework \cite{misra20213detr}.

\noindent \textbf{Image-guided Novel Object Selection.}
The geometrical correlation in a scene between 3D point clouds and paired 2D images provides a bridge to associate 2D detectors and 3D detectors. 
We first project the predicted 3D objects $\hat{\textbf{B}}$ to the 2D image coordinate. The projected boxes are denoted as $\hat{\textbf{b}}$ and can be obtained using camera parameters.
Then we select the novel objects with the guidance from 2D objects. 
For each 2D box $\textbf{b}_{m}$, we match it with the projected box $\hat{\textbf{b}}$ that has the highest overlapping area. We then filter out matched samples that have an overlap area less than a threshold, or those that are included in the base objects.
Let $\eta_{ij} = \eta(\textbf{b}_i, \textbf{b}_j )$ denote the Intersection over Union (IoU) \cite{misra20213detr} between two 2D boxes, we select novel objects as:
\begin{eqnarray} \label{eq:2d_filt}
\hat{\textbf{b}}^{\mathcal{N} }=\left \{ \hat{\textbf{b}}_{i} | \eta(\textbf{b}_{\text{m}}, \hat{\textbf{b}}_{i} ) \ge \max_{\eta_{mj}} \eta(\textbf{b}_{m}, 
\hat{\textbf{b}}_{j} ), \eta_{mi} \geq \epsilon, c_{m} \in C^{\mathcal{N} } \right  \}_{m=1}^M,
\end{eqnarray}
where $\epsilon$ is a threshold used to filter out cases with low IoU. 
We take the corresponding 3D objects from these selected 2D boxes $\hat{\textbf{b}}^{\mathcal{N} }$ and store them in the novel object memory bank. These novel 3D objects are subsequently used to retrain the 3D detector along with the base objects.
Periodically, we conduct the discovery procedure and update the entire memory bank with the newly discovered novel objects. 

\subsection{Hierarchical Cross-modal Feature Alignment}
\label{sec:hierarchical_align}
Pretrained large vision-language models (VLMs) have demonstrated remarkable success, showcasing powerful feature representation and generalization capabilities. Consequently, we align the 3D feature space with the pretrained vision-language model in a hierarchical design.
Specifically, this alignment takes place at three levels: instance, category, and scene levels. We will elaborate on each level below.

\vspace{-.01in}
\noindent \textbf{Instance-level 3D-Image Alignment.}
Building upon that image and point cloud have a natural correlation in geometry information, \eg shape, we directly associate 3D object features and pair 2D object features at instance level. 
Let $\textbf{f}^{3D}$ denote the 3D object feature, and $\textbf{f}^{2D}$ denote the corresponding cropped image feature generated from the VLM, \ie CLIP \cite{radford2021clip}.
We mitigate the distance between  $\textbf{f}^{3D}$ and  $\textbf{f}^{2D}$ by using L1-norm loss:
\begin{eqnarray}  \label{eq:loss-ins}
\mathcal{L}_{ins}^{\text{3d} \Rightarrow \text{rgb}} =  \left | \textbf{f}^{3D} - \textbf{f}^{2D} \right |.
\end{eqnarray}
The alignment at the instance level emphasizes the consistency in 3D features and image features, without yet considering general class information in the language domain.

\noindent \textbf{Class-level Cross-modal Alignment.}
We further align the 3D feature with the vision-language feature at the category level. Inspired by \cite{lu2023ov3det}, we categorize features from the three modalities by their class and use contrastive learning to bring together features of the same class while pushing apart those of different classes. 
In this arrangement, we use $\left \{ \textbf{g}_i \right \} _{i=1}^{S}$ to represent the set of $S$ features for all modalities (\ie point cloud, image and text) in a batch. The class labels for these features are denoted as $\left \{ c_i \right \} _{i=1}^{S}$.
We construct positive pairs by using samples from the same class and negative pairs by utilizing samples from different classes, then calculate the contrastive loss:
\begin{eqnarray} \label{eq:loss=cls}
\begin{split}
\mathcal{L}_{cls}^{3d \Rightarrow rgb,text}  = -\frac{1}{S}  \sum_{i=1}^{S} \log \frac{  {\textstyle \sum_{k=1}^{S}} \mathds{1} (i\neq k, c_i=c_k) \mathrm{e} ^{\textbf{g}_i \cdot \textbf{g}_k / \tau_1 }}{\sum_{j=1}^{S}  \mathrm{e} ^{\textbf{g}_i \cdot \textbf{g}_j / \tau_1 }}.
\end{split}
\end{eqnarray}
Here, $\tau_1$ is the temperature parameter, $\mathds{1}(\cdot)$ is the indicator function, which yields 1 when the condition is met and 0 otherwise.

\begin{figure}[t]
\centering
\includegraphics[width=0.6\textwidth]{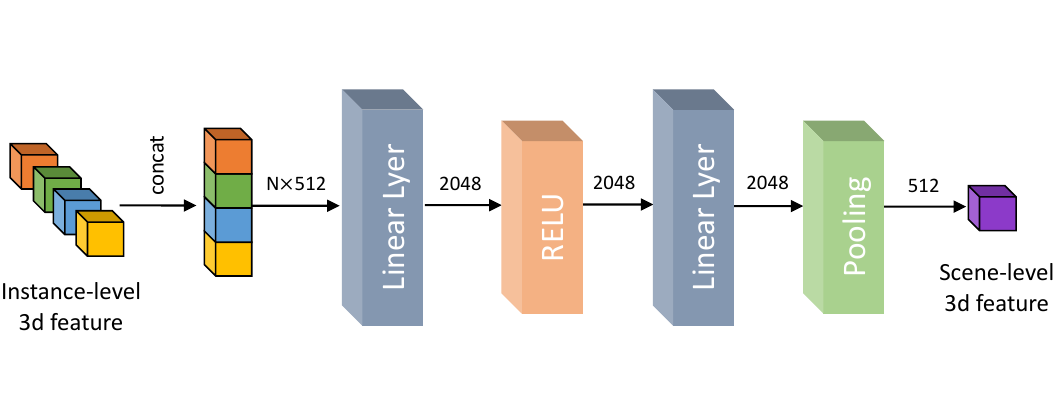} 
\vspace{-20pt}
\caption{\small{\textbf{Architecture of PISE.} The object features within a scene are concatenated and passed through two linear layers, sandwiched between a ReLU layer, to generate a high-dimensional feature. Subsequently, max pooling is applied to this high-dimensional feature to obtain a permutation-invariant scene-level feature.}
}
\label{fig:pise}
\vspace{-0.1in}
\end{figure}

\noindent \textbf{Scene-level Cross-modal Alignment.}
The objects within a scene often exhibit strong correlations, where certain objects are more likely to coexist. For example, in a living room, a bed is typically accompanied by a dresser, but not a refrigerator.
Utilizing this correlation prior is beneficial for aligning the cross-modal feature spaces at the scene level. Here, we align the 3D scene features with the text scene features. 
To generate the text scene feature, we first create a scene-level caption containing all class names present in the scene. This caption is then processed by the CLIP text encoder to produce the scene-level text feature $\textbf{z}^{text}$. 
For scene-level 3D feature extraction, we introduce the Permutation-Invariant Scene-level feature Extraction (PISE) module. This involves concatenating all individual 3D object features within a scene and projecting them into a high-dimensional space. To address permutation invariance, we utilize a max pooling operation on these high-dimensional features. The detailed architecture of the PISE module is illustrated in Figure \ref{fig:pise}.
With the scene-level 3D feature $\textbf{z}^{3D}$ extracted by the PISE module, we employ a contrastive loss to align the 3D scene features with the corresponding text scene features within a batch:
\begin{equation}
\mathcal{L}_{scene}^{3d \Rightarrow text} = -\frac{1}{L}  \sum_{i=1}^{L} \log \frac{ \mathrm{e} ^{ \textbf{z}^{3d}_{i} \cdot \textbf{z}^{text}_{i} / \tau_2 }}{\sum_{j=1}^{L}  \mathrm{e}^{\textbf{z}^{3d}_{i} \cdot\textbf{z}^{text}_{j} / \tau_2 }}.
\end{equation}
Here $L$ indicates the number of scenes in a batch, and $\tau_2$ is the temperature.

Considering all the alignment losses above, and along with the box regression loss, the total loss used in the third stage is:
\begin{eqnarray} \label{loss:align}
\begin{split}
    \mathcal{L}_{align} =  \mathcal{L}_{box}  +  \lambda_1 \mathcal{L}_{ins}^{3d\Rightarrow rgb} + \lambda_2 \mathcal{L}_{cls}^{3d\Rightarrow rgb,text}  + \lambda_3 \mathcal{L}_{scene}^{3d\Rightarrow text}.
\end{split}
\end{eqnarray}

\section{Experiment}
\subsection{Datasets}

Our proposed approach is evaluated on two challenging 3D indoor detection datasets: \textbf{SUN RGB-D}, which consists of 5,285 training samples with oriented 3D bounding box labels for 46 object categories. We select the 10 most frequent classes as base classes and the remaining 36 
classes as novel classes; \textbf{ScanNetv2} \cite{dai2017scannet}, it has 1,201 training samples with 200 object categories. We use the 10 most frequent classes as base classes and the remaining 50 most frequent classes as novel classes. All configurations above are the same as in \cite{cao2023coda}.

Additionally, we adapt the settings in OV3DET \cite{lu2023ov3det}, using generated pseudo-labels trained on all classes. However, L3DET \cite{zhu2023l3det} has a different configuration compared to \cite{lu2023ov3det}, where it uses 10 classes for seen class training and another 10 classes for unseen class. For a fair comparison, we select the 10 overlapping novel classes (toilet, bed, chair, sofa, dresser, table, cabinet, bookshelf, pillow, and sink) between \cite{lu2023ov3det} and the novel classes in \cite{zhu2023l3det} for validation. We denote this benchmark as \textbf{ScanNet-10}.

\subsection{Baselines and Evaluation Metrics}
\noindent \textbf{Baselines.}
For benchmark datasets SUN RGB-D and ScanNet-10, we select Det-PointCLIP \cite{zhang2022pointclip}, Det-PointCLIPv2 \cite{zhu2022pointclip2}, Det-CLIP \cite{zeng2023clip2}, 3D-CLIP \cite{radford2021clip}, and CoDA \cite{cao2023coda} as our comparative methods \cite{cao2023coda}. For the ScanNet-10 benchmark, we compared our method with L3DET \cite{zhu2023l3det}, OV-3DET \cite{lu2023ov3det}, and CoDA \cite{cao2023coda}.

\noindent \textbf{Metrics.}
Regarding validation metrics, we employ mean Average Precision (mAP) and mean Average Recall (mAR) \cite{misra20213detr}, with an IoU threshold set to 0.25. Among these two metrics, mAP is our primary metric.

\subsection{Implementation Details} \label{sec:implemention}
\noindent \textbf{Training Strategy and Hyperparameters.}
Our training procedure includes three stages.
In the first stage, \ie base detector training, we train the 3D detector for 1000 epochs for a fair comparison with \cite{cao2023coda}. In the second stage, \ie novel class discovery, we train for an additional 200 epochs. In the final stage, \ie feature space alignment, we train another 50 epochs to align the 3D feature space to the vision-language feature space.
For the first stage, we configure a batch size of 8 and a learning rate of 0.0004, utilizing 128 queries. Subsequently, during the second and third stages, we specifically adjust the learning rate to 0.0001, batch size to 16, and increase the query size to 196 for enhanced detector training and feature alignment.
We set the temperatures $\tau_1$ and $\tau_2$ to 1.0 and the threshold $\epsilon$ to 0.75. Additionally, the hyperparameters $\lambda_1$, $\lambda_2$, and $\lambda_3$ are initially set to 0.02 during warm-up and later adjusted to 1, 1, and 0.5, respectively.

\noindent \textbf{Model Selection and Prompt Setting.}
We modify 3DETR \cite{misra20213detr} as our 3D detector by removing the classification head and using only the bounding box regression loss. The pretrained Detic \cite{zhou2022detic} is utilized as our 2D detector. For feature alignment, we leverage the pretrained CLIP \cite{radford2021clip} to encode image and text features.
In the class-level and scene-level feature space alignment, we generate text prompts for language feature encoding. Specifically, for the class-level prompt, we generate feature embeddings using the template “\textit{A photo of} [class name].”, where [class name] represents the category name.
For the scene-level prompt, we concatenate the class names in a scene into a list and insert the list into the template “\textit{A room with} [class list].”, where [class list] denotes the position to insert the list. Additionally, we include the non-object description prompt, \ie, “\textit{A photo of nothing}.” to represent areas without discriminative objects.

\begin{table*}[t] 
\centering
\caption{\small{Comparison results(\%) of our INHA and baseline methods on SUN RGB-D. The label “Need image” signifies that image input is required for inference.}}
\rowcolors{3}{grey!8}{blue!0} 
\resizebox{1\hsize}{!}{
\begin{tabular}{l||c||cc|c||cc|c}
\toprule
Method & Need image &  $\mathrm{mAP}_{\mathrm{Novel}}$ &  $\mathrm{mAP}_{\mathrm{Base}}$ & \textbf{Avg.}  &  $\mathrm{mAR}_{\mathrm{Novel}}$ &  $\mathrm{mAR}_{\mathrm{Base}}$ & \textbf{Avg.} \\
\midrule
Det-PointCLIP	\cite{zhang2022pointclip} & $\times$  &0.09	&5.04	&1.17	&21.98	&65.03	&31.33   \\
Det-PointCLIPv2 \cite{zhu2022pointclip2}  & $\times$	 &0.12	&4.82	&1.14	&21.33	&63.74	&30.55  \\
Det-CLIP 	\cite{zeng2023clip2} & $\times$ &0.88	&22.74	&5.63	&22.21	&65.04	&31.52 \\
3D-CLIP    \cite{radford2021clip}  & \checkmark	&3.61	&30.56	&9.47	&21.47	&63.74	&30.66     \\
CoDA \cite{cao2023coda}   & $\times$	&6.71	&38.72	&13.66	&33.66	&66.42	&40.78   \\
\midrule
\textbf{INHA (Ours)}    & $\times$	& \textbf{8.91}	&\textbf{42.17}	&\textbf{16.18}	&\textbf{51.34}	&\textbf{78.65}	&\textbf{57.23 }    \\
\bottomrule
\end{tabular}}
\label{tb:result_sunrgbd}
\end{table*}

\begin{table*}[t] 
\centering
\caption{\small{Comparison results(\%) of our INHA and baseline methods on ScanNetv2.}}
\rowcolors{3}{grey!8}{blue!0} 
\resizebox{1\hsize}{!}{
\begin{tabular}{l||c||cc|c||cc|c}
\toprule
Method & Need image &  $\mathrm{mAP}_{\mathrm{Novel}}$ &  $\mathrm{mAP}_{\mathrm{Base}}$ & \textbf{Avg.}  &  $\mathrm{mAR}_{\mathrm{Novel}}$ &  $\mathrm{mAR}_{\mathrm{Base}}$ & \textbf{Avg.} \\
\midrule
Det-PointCLIP	\cite{zhang2022pointclip}  & $\times$& 0.13	& 2.38	& 0.5	& 33.38	& 54.88	& 36.96 \\
Det-PointCLIPv2	\cite{zhu2022pointclip2} & $\times$& 0.13	& 1.75	& 0.4	& 32.6	& 54.52	& 36.25 \\
Det-CLIP2	\cite{zeng2023clip2} & $\times$& 0.14	& 1.76	& 0.4	& 34.26	& 56.22	& 37.92 \\
3D-CLIP	   \cite{radford2021clip}  & \checkmark & 3.74 & 14.14	& 5.47	& 32.15	& 54.15	& 35.81 \\
CoDA	 \cite{cao2023coda} & $\times$ & 6.54 & 21.57	& 9.04	& 43.36	& 61.0	& 46.3 \\
\midrule
\textbf{INHA (Ours)}  & $\times$   &	 \textbf{7.79}	&  \textbf{25.1}	&  \textbf{10.68}	&  \textbf{55.1}	&  \textbf{71.6}	&  \textbf{57.85} \\
\bottomrule
\end{tabular}
}
\label{tb:result_scannet}
\end{table*}

\subsection{Main Results}

\noindent \textbf{Results on SUN RGB-D.}
We evaluate our method and baseline methods on the SUN RGB-D dataset, and the results are shown in Table \ref{tb:result_sunrgbd}. As can be seen, our method outperforms all other methods on both base class and novel class in terms of both mean average precision and mean average recall. Specifically, compared to the state-of-the-art method CoDA, our method has a 30\% higher performance on $\text{mAP}_{\mathrm{Novel}}$ and a 10\% higher performance on the base class $\text{mAP}_{\mathrm{Base}}$. This highlights that our method not only finds more novel objects but also better aligns the 3D feature space to the vision-language feature space, yielding superior performance.

\begin{table*}[t]
\centering
\caption{\small{Comparison results(\%) of our INHA and baseline methods on ScanNet-10.}}
\rowcolors{3}{grey!8}{blue!0} 
\resizebox{1\hsize}{!}{
\begin{tabular}{l||cccccccccc|c}
\toprule
Method & toilet & bed & chair & sofa & dresser & table & cabinet & bookshelf & pillow & sink & \textbf{Avg. }\\
\midrule
L3DET \cite{zhu2023l3det} & 56.34 & 36.15 & 16.12 & 23.02 & 8.13 & 23.12 & \textbf{14.73} & \textbf{17.27} & 23.44 & 27.94 & 24.63 \\
OV-3DET \cite{lu2023ov3det} & 57.29 & 42.26 & 27.06 & 31.5 & 8.21 & 14.17 & 2.98 & 5.56 & 23 & 31.6 & 24.36 \\
CoDA \cite{cao2023coda} & \textbf{68.09} & 44.04 & 28.72 & \textbf{44.57} & 3.41 & 20.23 & 5.32 & 0.03 & \textbf{27.95} & \textbf{45.26} & 28.76 \\
\midrule
\textbf{INHA (Ours)}  & 67.4 & \textbf{46.01} & \textbf{33.32} & 40.92 & \textbf{9.1} & \textbf{26.42} & 4.28 & 11.3 & 26.15 & 35.69 & \textbf{30.06} \\
\bottomrule
\end{tabular}
}
\label{tab:scannet-10}
\end{table*}

\noindent \textbf{Results on ScanNetv2.}
We also evaluate our method and baseline methods on the ScanNetv2 dataset, and the results are shown in Table \ref{tb:result_scannet}. As can be seen, our method continually outperforms other methods on both base class and novel class. We have a 19\% higher performance on novel class $\text{mAP}_{\mathrm{Novel}}$ and a 16.4\% higher performance on base class $\text{mAP}_{\mathrm{Base}}$. All the results highlight the significant novel class discovery of our method.

\subsection{Results on ScanNet-10}
As OV3DET \cite{lu2023ov3det} leverages a 2D detector to generate pseudo labels for all classes, we adopt this pseudo-labeling setting for comparison. We evaluate our methods alongside several others on the ScanNet-10 benchmark, and the results are presented in Table \ref{tab:scannet-10}. 
Notably, L3DET \cite{zhu2023l3det} employs synthetic data to expand the novel category on top of the 10 base classes. To ensure a fair comparison between \cite{lu2023ov3det} and \cite{zhu2023l3det}, we select the overlapping 10 novel classes.
From the results, it is evident that our method outperforms the others and achieves the best performance in terms of mean average precision.

\begin{table*}[t]
\centering
\caption{\small{\textbf{Component study on SUN RGB-D.} We evaluate the main components of our proposed method, starting with the base method that utilizes only base class box labels to train a locator, then we incrementally incorporate key components of our proposed method.}
}
\rowcolors{2}{blue!0} {grey!8}
\resizebox{0.9\textwidth}{!}{%
\begin{tabular}{c||ccc||cc|cc}
\toprule
IGND  & Instance-level & Class-level & Scene-level & $\mathrm{mAR}_{\mathrm{Novel}}$ & $\mathrm{mAR}_{Base}$ & $\mathrm{mAP}_{\mathrm{Novel}}$ & $\mathrm{mAP}_{\mathrm{Base}}$ \\
\midrule
  &  &  &  & 34.12 & 77.36 & 0.19 & 4.03\\
\midrule
\checkmark &  &  &  & 51.12 & 78.31 & 0.35 & 2.07 \\
\midrule
\checkmark & \checkmark  &  &  & 50.70 & \textbf{79.11} & 6.85 & 38.46 \\
\checkmark &  \checkmark &  \checkmark &  & 50.75 & 78.27 & 8.03 & 40.21\\
\checkmark &  \checkmark &  \checkmark &  \checkmark & \textbf{51.34} & 78.65 & \textbf{8.91} & \textbf{42.17} \\
\bottomrule
 \end{tabular}
 }
\label{tbl:abla}
\vspace{-0.1in}
\end{table*}

\subsection{Ablation Study}

\begin{wrapfigure}{r}{0.5\textwidth}
\vspace{-0.25in}
\centering
\vspace{-0.2in}
\includegraphics[width=0.5\textwidth]{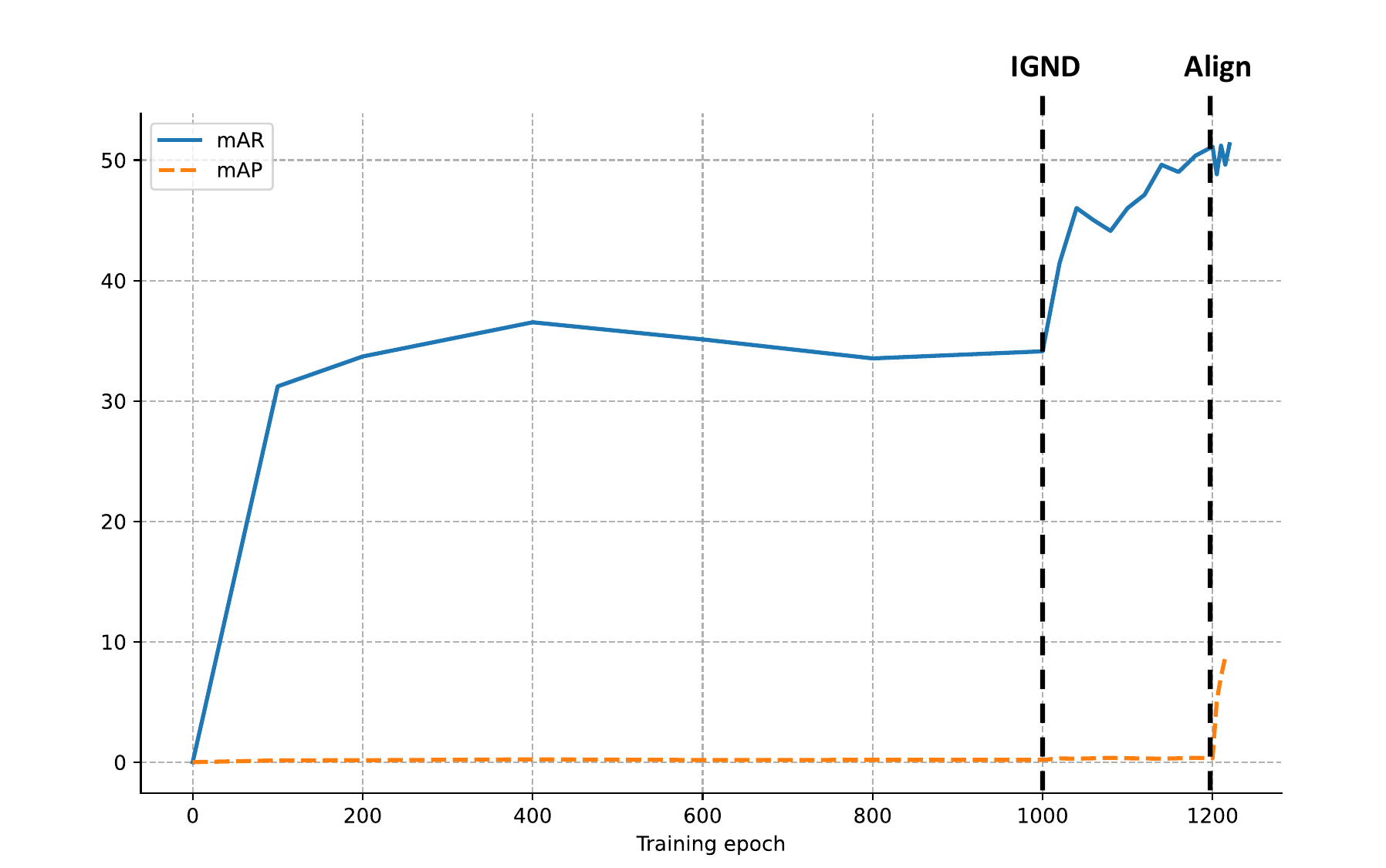}
\caption{\small{Mean average recall (mAR) and mean average precision (mAP) for novel classes were tracked during training epochs on SUN RGB-D. } 
}
\label{img:training_iter}
\vspace{-0.3in}
\end{wrapfigure}

\noindent \textbf{Effect of Image-Guided Novel Class Discovery.}
To assess the effectiveness of our proposed components, we conduct an ablation experiment on the SUN RGB-D dataset, evaluating various versions of our method. The results are summarized in Table \ref{tbl:abla}. The inclusion of the IGND module leads to a significant improvement in novel object discovery, resulting in higher performance on the mean average recall ($\text{mAR}_{\mathrm{Novel}}$) for novel classes.
We visualize the changes in mean average recall and mean average precision throughout the training process, as depicted in Figure \ref{img:training_iter}. It is evident that with the introduction of IGND, the mean average recall of novel classes notably increases (specifically at the epoch marked by \textit{IGND}), further underscoring the effectiveness of the image-guided novel object discovery module.

\noindent \textbf{Effect of Hierarchical Cross-modal Feature Alignment.}
When incorporating instance-level, class-level, and scene-level feature space alignment, respectively, the mean average precision (including both $\text{mAP}_{\mathrm{Novel}}$ and $\text{mAP}_{\mathrm{Base}}$) demonstrates a gradual improvement, as illustrated in Table \ref{tbl:abla}. Moreover, as depicted in Figure \ref{img:training_iter}, the implementation of our hierarchical multi-modality feature space alignment (initiated at the epoch beginning with \textit{Align}) leads to a significant enhancement in the mean average precision on novel classes. These findings underscore the effectiveness of our proposed hierarchical design in aligning the feature spaces. 

\begin{wraptable}{r}{0.5\textwidth}
\vspace{-0.45in}
\centering
\caption{\small{Mean average recall ($\text{mAR}_{\mathrm{Novel}}$) for novel classes w and w/o Image-Guided Query Seed Initialization in the IGND stage. 
In the absence of Image-Guided Query Seed Initialization, the farthest point sampling algorithm \cite{qi2017pointnetplusplus} is utilized for additional seed initialization.}
}
\resizebox{0.5\textwidth}{!}{
\begin{tabular}{c|c|c}
\toprule
Image-guided Novel  & Image-guided Query & \multirow{2}{*}{$\mathrm{mAR}_{\mathrm{Novel}}$} \\
 Object Selection & Seed Initialization &   \\
\midrule
\checkmark &  & 48.31 \\
\checkmark & \checkmark  & \textbf{51.12} \\
\bottomrule
\end{tabular}
}
\label{tbl:2d_query}
\vspace{-0.3in}
\end{wraptable}

\noindent \textbf{Effect of Image-guided Query Seed Initialization.}
To evaluate the efficacy of the Image-Guided Query Seed module, we conduct experiments wherein the module is excluded while maintaining a constant query size of 196 on SUN RGB-D. The results are presented in Table \ref{tbl:2d_query}. Notably, the results indicate that the absence of guided query seeds adversely impacts the mean average recall for Novel classes.
These outcomes underscore the importance of integrating additional query seeds derived from 2D object centers. Such integration enables the model to identify more objects, thereby emphasizing the effectiveness of leveraging vision models to enhance 3D detection capabilities.

\subsection{Visualization}

\begin{figure}[t]
\centering
\includegraphics[width=1\textwidth]{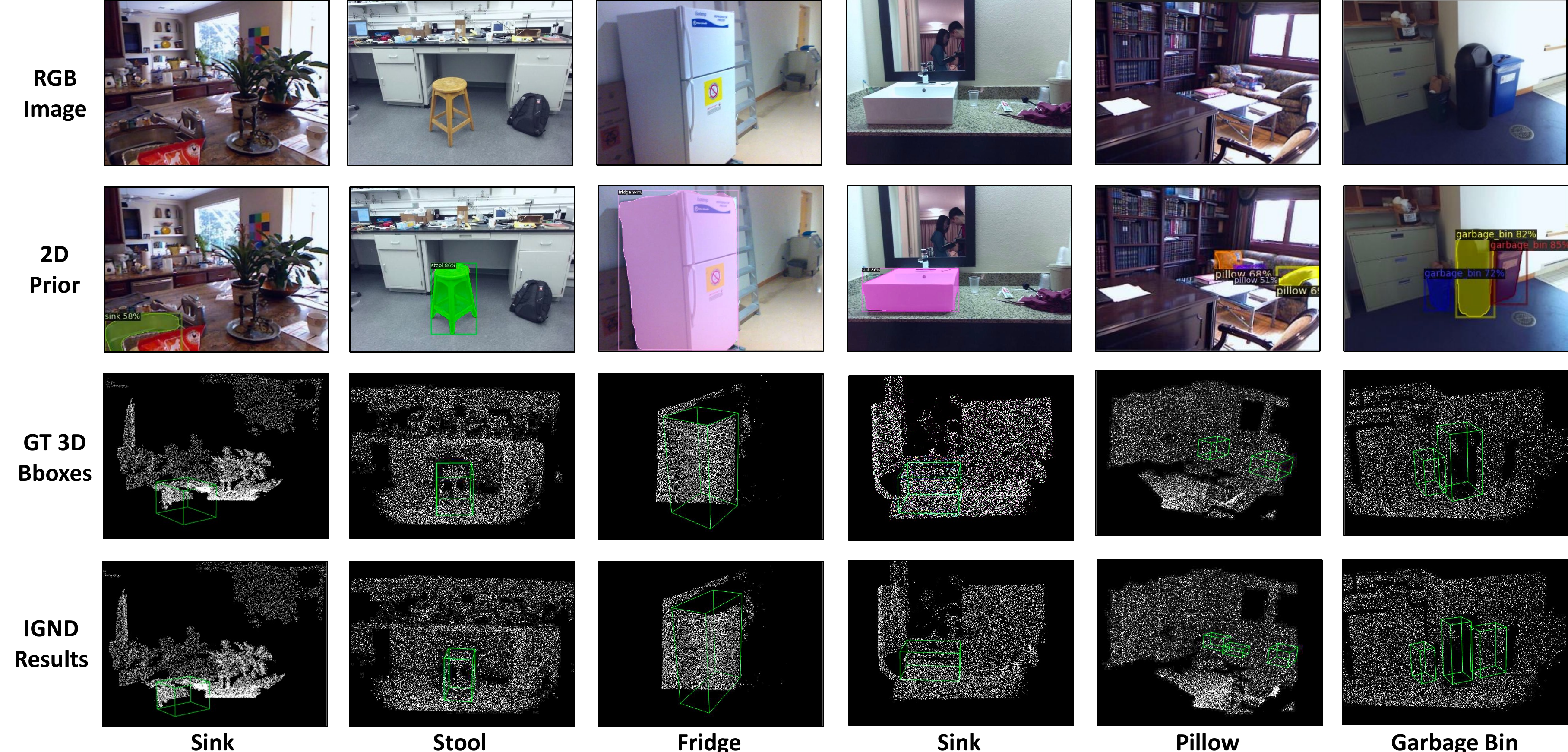} 
\vspace{-0.2in}
\caption{\small{\textbf{Quality results of IGND.} From top to down, the sequence includes the original RGB image, the detected 2D objects from the 2D detector, the ground truth 3D objects, and the discovered novel objects from the IGND module.}
}
\label{img:quality_result}
\vspace{-0.1in}
\end{figure}

\noindent \textbf{Qualitative Results of IGND.}
Our proposed Image-Guided Novel Class Discovery (IGND) effectively discovers novel classes, as evidenced by the high-quality results depicted in Figure \ref{img:quality_result}. In the 2$^{nd}$ row of the figure, the fabulous results in 2D object detection from the open-vocabulary 2D detector, \ie, Detic \cite{zhou2022detic}, showcase the power of the vision foundation model.
By effectively leveraging 2D objects detected by the vision foundation model, our IGND accurately identifies novel classes, even including objects not annotated in the ground truth. For example, in the 6$^{th}$ column of Figure \ref{img:quality_result}, IGND discovers a garbage bin that was not included in the ground truth annotations.
These high-quality qualitative results underscore the effectiveness of our proposed IGND in discovering more reliable 3D novel objects.

\begin{figure*}[t]
\centering
\includegraphics[width=0.88\textwidth]{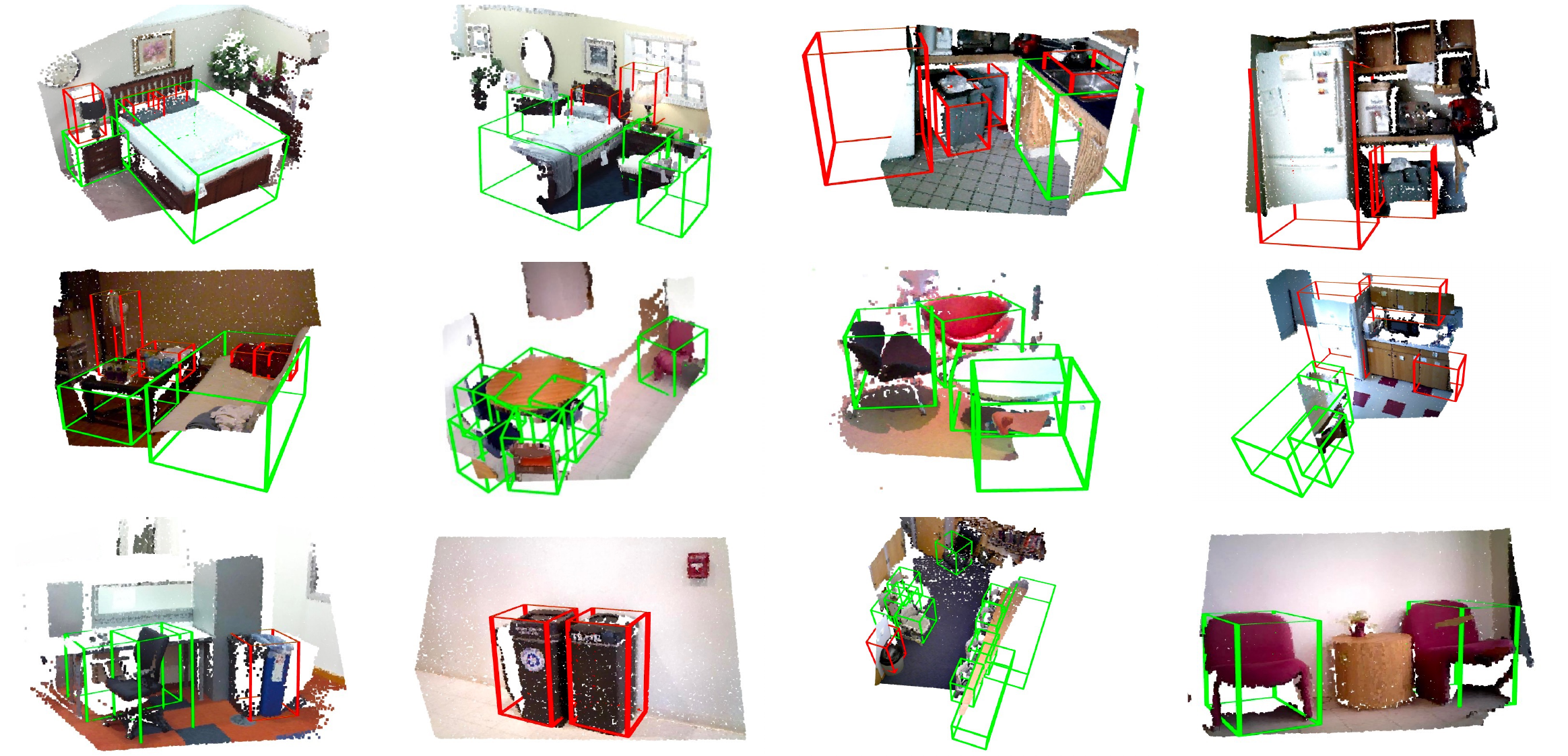} 
\vspace{-0.1in}
\caption{
\textbf{Quality Results of INHA on OV-3DDet.} The base objects are indicated by green boxes, while the novel objects are indicated by red boxes. The base classes include bed, table, sofa, chair, toilet, desk, dresser, nightstand, bookshelf, and bathtub.
}
\label{img:qr}
\vspace{-0.1in}
\end{figure*}

\noindent \textbf{Qualitative Results of INHA.}
In addition, we present visualizations of the high-quality results achieved by INHA in open-vocabulary 3D object detection. Exemplary outcomes obtained from the SUN RGB-D dataset are illustrated in Figure \ref{img:qr}. These visualizations serve as evidence of the impressive open-vocabulary capabilities demonstrated by our INHA exhibit in these cases.

\section{Conclusion}
In this study, we delve into the challenging realm of open-vocabulary 3D object detection.
While vision language foundation models have propelled 2D detection forward in open-vocabulary scenarios, their potential for 3D detection remains underutilized. To address this gap, we introduce a novel framework: the image-guided novel class discovery and hierarchical feature space alignment framework, dubbed as INHA. 
Specifically, we integrate 2D detection model and 3D detector to discover novel objects. Additionally, we hierarchically align the 3D features with the vision-language feature space at the instance, category, and scene levels.
Our INHA capitalizes the power of foundation models to extract comprehensive guidance information from both text and images, which is then effectively integrated with 3D inputs for open-vocabulary 3D object detection. 
Extensive experiments validate the effectiveness of our INHA across various datasets.

\vspace{0.05in}
\noindent \textbf{Acknowledgments}
This research work is partially supported by the Shanghai Science and Technology Program (Project No. 21JC1400600), and the Agency for Science, Technology and Research (A*STAR) under its MTC Programmatic Funds (Grant No. M23L7b0021).

\clearpage  

%
%
\bibliographystyle{splncs04}
\bibliography{main}

\end{document}